\DeclareUnicodeCharacter{0301}{\'e}

\documentclass[journal,article,submit,pdftex,moreauthors]{mdpi} 
\firstpage{1} 
\pubvolume{1}
\issuenum{1}
\articlenumber{0}
\pubyear{2025}
\copyrightyear{2025}
\datereceived{ } 
\daterevised{ } %
\dateaccepted{ } 
\datepublished{ } 
\hreflink{https://doi.org/} %

\definecolor{customgreen}{RGB}{0, 90, 0}
\definecolor{customhighgreen}{RGB}{0, 120, 0}
\Title{Evolutionary Optimization for the Classification of Small Molecules Regulating the Circadian Rhythm Period: A Reliable Assessment}

\TitleCitation{Evolutionary Optimization for the Classification of Small Molecules Regulating the Circadian Rhythm Period: A Reliable Assessment}

\Author{Antonio Arauzo-Azofra $^{1}$ \orcidA{}, Jose Molina-Baena $^{2}$ \orcidB{}, and Maria Luque-Rodriguez $^{2,}$* \orcidC{}} 

\AuthorNames{Antonio Arauzo-Azofra, Jose Molina-Baena and Maria Luque-Rodriguez}

\isAPAStyle{%
       \AuthorCitation{Arauzo-Azofra, A., Molina-Baena, J., \& Luque-Rodriguez, M.}
         }{%
        \isChicagoStyle{%
        \AuthorCitation{Arauzo-Azofra, A., Molina-Baena, J., \& Luque-Rodriguez, M.}
        }{
        \AuthorCitation{Arauzo-Azofra, A., Molina-Baena, J., \& Luque-Rodriguez, M.}
        }
}

\address{%
$^{1}$ \quad Dept. of Project Engineering, Universidad de Córdoba; arauzo@uco.es\\
$^{2}$ \quad Dept. of Computer Science and Artificial Ingelligence, Universidad de Córdoba; i42mobaj@uco.es}

\corres{Correspondence: mluque@uco.es}

\abstract{The circadian rhythm plays a crucial role in regulating biological processes, and its disruption is linked to various health issues. Identifying small molecules that influence the circadian period is essential for developing targeted therapies. This study explores the use of evolutionary optimization techniques to enhance the classification of these molecules. We applied an evolutionary algorithm to optimize feature selection and classification performance. Several machine learning classifiers were employed, and performance was evaluated using accuracy and generalization ability. The findings demonstrate that the proposed evolutionary optimization method improves classification accuracy and reduces overfitting compared to baseline models. Additionally, the use of variance in accuracy as a penalty factor may enhance the model’s reliability for real-world applications. Our study confirms that evolutionary optimization is an effective strategy for classifying small molecules regulating the circadian rhythm. The proposed approach not only improves predictive performance but also ensures a more robust model. Future work will focus on refining the genetic algorithm.}


\begin{document}

\setcounter{section}{-1} %

\section{Introduction}

The circadian rhythm regulates biological processes in 24-hour cycles, and its alteration is associated with sleep disorders and can be related with pathologies such as cancer~\citep{jagielo2023circadian}. The circadian clock plays a fundamental role in the regulation of sleep, and its stability is increasingly recognized as a determining factor in various biological processes, which are essential for maintaining good health. To this end, photopharmacological manipulation of a core clock protein, mammalian CRY1, is an effective strategy for regulation of the circadian clock. Supported by the analysis of data and advanced computational modelling techniques, it has established as an essential tool in the detection and characterization of compounds with an impact on circadian dynamics~\citep{kolarski2021photopharmacological}. Recent studies apply high-throughput screening and data analysis to identify potential drugs, taking advantage of advanced modelling and machine learning techniques~\cite{wildey2017high}. 

Structural approaches in drug discovery optimize efficiency and reduce costs, while the incorporation of screening methods allows the elimination of inappropriate molecules, such as toxic or inactive ones. After identifying 171 molecules that target functional domains of CRY1 protein, using structure-based drug design methods, and experimentally determining that 115 of these molecules were nontoxic, Gul et al.~\citep{gul2021structure} performed a machine learning study to classify molecules by identifying features that make them toxic. They also addressed the classification of the same molecules based on their effect on CRY1. While both problems are considered challenging, only the second has been further investigated in other studies~\cite{chawathe2022attribute}.

The problem of learning to determine toxicity from these 171 molecules is complex, characterized by many features and few examples. As a result, it is considered ill-posed, as traditional statistical methods often struggle with insufficient data. In this context, overfitting becomes a significant risk, as models can easily memorize the training data instead of generalizing. Therefore, we classify it as an overfitting-prone problem. Tackling these handicaps constitutes an interesting challenge in machine learning~\citep{kuncheva2020feature}. These handicaps motivate us to apply improved automated feature selection while remaining fully aware of its limitations.

In this paper, we reproduce the machine learning experimentation from~\citep{gul2021structure}  revealing its weaknesses and evaluate the use of a more advanced feature selection process based on genetic algorithm meta-heuristic search to achieve improved and more trustworthy results.

\section{Genetic algorithm for feature selection}

Evolutionary algorithms, inspired by natural evolution, are designed to optimize solutions. Among them, genetic algorithms (GAs) are particularly well-suited for feature selection (FS). These algorithms start with a population of individuals, each representing a possible solution. Through processes of selection, crossover and mutation, successive generations evolve towards better solutions (Figure ~\ref{fig:genetic}). The following subsections describe the main aspects of the algorithm used in this work.

\subsection{Evolution}
The framework of the algorithm is based on the evolution of a single population of solutions over multiple generations.

\begin{figure}[htbp]
  \centering
  \includegraphics[width=0.8\textwidth]{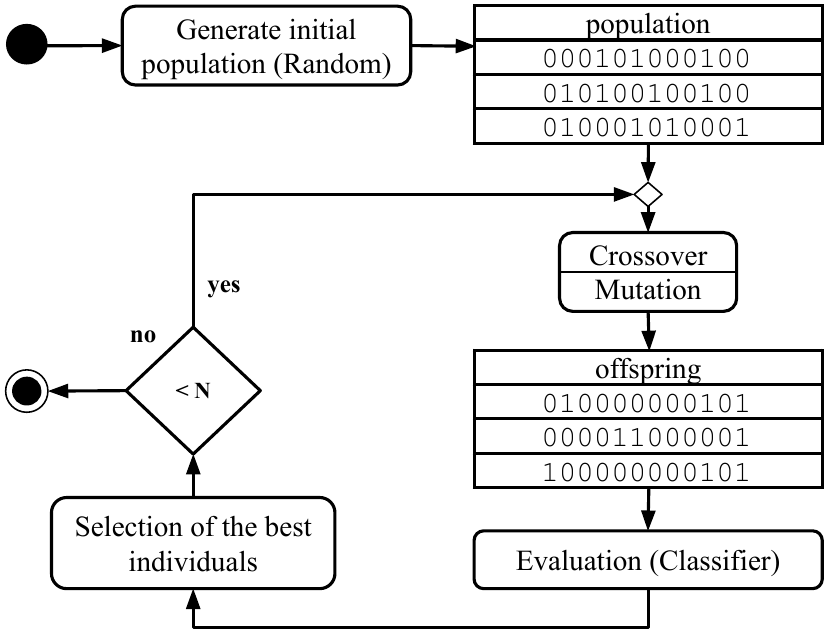}
\caption{Genetic algorithm activity diagram.}
\label{fig:genetic}
\end{figure}

Once the population has been initialized, each evolutionary cycle begins with the application of crossover and mutation operators to generate new solutions. Subsequently, the individuals in the population are evaluated and selected, based on their performance in the target task, to form the next generation with the objective of progressively improving the quality of the solutions (Figure ~\ref{fig:genetic}).

Figure \ref{fig:genetic} illustrates the main steps of the algorithm.

\subsection{Encoding}
\label{sec:encoding}
The encoding of solutions determines how potential solutions to the problem are represented. Each solution is encoded as a genotype, which serves as an abstract representation of the solution. To evaluate its quality, the genotype is decoded into its corresponding phenotype, representing the actual interpretation within the problem domain. The effectiveness of the encoding scheme directly influences the algorithm's ability to explore and optimize solutions efficiently.

In this work, the chosen encoding scheme is binary. Each individual in the population represents a possible solution to the problem, which is described as a vector of length n. Each value in the vector corresponds to a feature of the problem, with a 1 indicating that the feature is selected and a 0 otherwise.

\subsection{Initialization}
In an evolutionary algorithm, the initial generation of the population is generated randomly, assigning to each individual an initial configuration that represents a possible solution to the problem. However, the nature of the problem significantly influences the optimal configuration of individuals. Depending on the dimensionality of the data set and the complexity of the problem, the expected number of selected features may vary considerably.

GAs typically initialize individuals with a uniform probability of inclusion or exclusion of each feature, resulting in an expected selection of 50\% of the features. However, in high dimensionality datasets, this strategy can generate over-fitted initial solutions, which significantly slows down the evaluation of individuals. Since the GA will naturally adjust the number of features selected based on their impact on performance, a more efficient initialization strategy may be to start with a smaller number of features when working with high-dimensional spaces~\citep{feng2024feature}.

To strike a balance between simplicity and generality, the initialization strategy used in our algorithm is based on an adjustable likelihood as parameter~\citep{luque2022initialization}.

\subsection{Crossover operator}

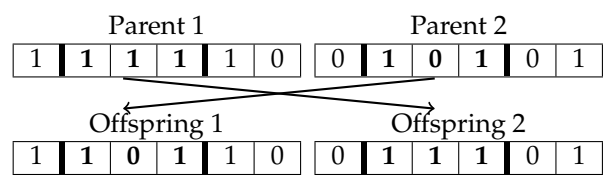
\begin{figure}[htbp]
\begin{tikzpicture}[overlay, remember picture]
    \draw[black, thick, ->] (4.5, -0.92) -- (8.6, -1.32);
    \draw[black, thick, ->] (8.6, -0.92) -- (4.5, -1.32); 
\end{tikzpicture}

\centering
\begin{tabular}{| c !{\vrule width 2pt} c | c | c !{\vrule width 2pt} c | c |}
\multicolumn{6}{ c }{Parent 1} \\ \hline
1 & \textbf{1} & \textbf{1} & \textbf{1}  & 1 & 0\\ \hline
\multicolumn{6}{c}{}\\
\multicolumn{6}{ c }{Offspring 1} \\ \hline
1 & \textbf{1} & \textbf{0}  & \textbf{1}  & 1 & 0\\ \hline
\end{tabular}
$\overset{}{}$ 
\begin{tabular}{| c !{\vrule width 2pt} c | c | c !{\vrule width 2pt} c | c |}
\multicolumn{6}{ c }{Parent 2} \\ \hline
0 & \textbf{1} & \textbf{0}  & \textbf{1}  & 0 & 1 \\ \hline
\multicolumn{6}{c}{}\\
\multicolumn{6}{ c }{Offspring 2} \\ \hline
0 & \textbf{1} & \textbf{1}  & \textbf{1}  & 0 & 1 \\ \hline
\end{tabular}
\caption{Crossover operator.}
\label{fig:cruce}
\end{figure}

The crossover operation combines coded individuals to explore potentially better solutions. In this work, two-point crossover, a method widely used in GAs, is employed to generate the next generation within the population. This mechanism exchanges a segment of the genotypic vector between two parent individuals, bounded by two randomly selected points. Figure~\ref{fig:cruce} illustrates this process. The probability of selecting segments of different sizes and their location within the genotype is uniform, ensuring a balanced exploration of the search space without introducing biases in the optimization.

\subsection{Mutation operator}

\begin{figure}[htbp]
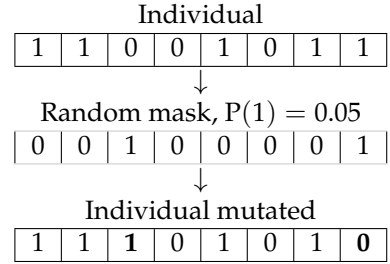

\centering
\begin{tabular}{| c | c | c | c | c | c | c | c |}
\multicolumn{8}{c}{Individual} \\ \hline
1 & 1 & 0 & 0  & 1 & 0 & 1 & 1\\ \hline
\multicolumn{8}{c}{$\downarrow$}\\
\multicolumn{8}{c}{Random mask, $\text{P}(1)=0.05$} \\ \arrayrulecolor[gray]{0.8} \hline
0 & 0 & 1 & 0  & 0 & 0 & 0 & 1\\ \hline \arrayrulecolor{black}
\multicolumn{8}{c}{$\downarrow$}\\
\multicolumn{8}{ c }{Individual mutated} \\ \hline
1 & 1 & \textbf{1}  & 0 & 1 & 0 & 1 & \textbf{0}\\ \hline
\end{tabular}
\caption{Mutation operator.}
\label{fig:mutacion}
\end{figure}

The mutation operation on encoded individuals introduces random modifications in the population to preserve genetic diversity. This mechanism is crucial for preventing premature convergence, allowing the algorithm to explore a broader search space and reducing the risk of getting trapped in local optima. In this study, the bit-flip mutation operator is employed, a widely used technique in binary-encoded GAs. This operator selectively alters certain binary values within the genotype of each individual, flipping them with a predefined probability, which serves as a control parameter for the mutation intensity. Figure \ref{fig:mutacion} illustrates this process.

\subsection{Fitness evaluation}
Fitness evaluation in a GA is the process of assessing how well each individual in the population solves the given problem. The fitness function quantitatively measures the quality of each solution based on a predefined criterion. In the case of feature selection problems, this criterion is the efficiency of the selected feature subset. The fitness score determines the likelihood of an individual being selected for reproduction, guiding the evolutionary process toward optimal solutions over successive generations.

In this work, we use the wrapper approach with 10-fold cross-validation to evaluate the quality of each individual (feature subset). The fitness of an individual is determined by training a machine learning model using the selected features, represented by the genotype, and evaluating its performance based on the accuracy metric. Furthermore, to encourage compact feature subsets, we include a penalty term to balance performance and feature reduction (Equation ~\ref{eq:fitness}).
\begin{equation}
\label{eq:fitness}
    \text{Fitness} = (1 - \alpha) \cdot \text{Effectiveness} + \alpha \cdot \frac{|F| -|S|}{|F|}\\
\end{equation}

where $\alpha$ is the weight parameter, $|S|$ is the number of selected features and $|F|$ is the total number of features in the dataset, and \text{Effectiveness} is the the average accuracy across all test folds

To mitigate over-fitting, we define a second fitness function (Equation ~\ref{eq:fitness-variance}), which incorporates the variance of accuracy as an additional penalty factor. The smaller the variance in accuracy, the more stable the model’s behavior, meaning it performs consistently across all test runs. A lower variance indicates that the model generalizes well, reducing the likelihood of over-fitting. This stability suggests that the algorithm is not overly dependent on specific training data patterns. Our hypothesis is that this will make it a better generalizer, more reliable for real-world applications.

\begin{equation}
\label{eq:fitness-variance}
    \text{Fitness} = (1 - \alpha) \cdot (\text{Effectiveness} - \text{variance}) + \alpha \cdot \frac{|F| -|S|}{|F|}\\
\end{equation}

\subsection{Selection}
Selection in a GA is the process of choosing individuals from the current population to create offspring for the next generation. The selection method directly influences the convergence and performance of the algorithm by favoring individuals with higher fitness scores while maintaining genetic diversity. The goal of selection is to balance exploration and exploitation, ensuring that the algorithm effectively searches the solution space while avoiding premature convergence.

In this work, we employ a binary tournament selection approach, where two individuals are randomly selected from the population, and the one with the higher fitness value is chosen as a parent for the next generation. This process is repeated until the required number of parents is selected. Binary tournament selection balances exploration and exploitation of the search space, allowing the most promising solutions to propagate across generations while preserving genetic diversity.

\section{Material and Methods}
This section describes the experimental methodology used to reproduce the previous experimentation and to evaluate the performance of feature selection with a GA meta-heuristic search on the Toxicity classification problem. 
In this study, we aim to validate three main hypotheses. First, we hypothesize that the proposed validation method will demonstrate the presence of overfitting in the results reported by the previous study. Second, we aim to improve the results of the compared approach by leveraging an GA. Third, the use of variance of accuracy as penalty factor will make it a better generalizer.
The experiments were designed to assess the effectiveness of our approach under different conditions. We provide details on the dataset characteristics, the classifiers used in the experiments, the execution environment, and the parameters adopted in the GA.

\subsection{Dataset}
The dataset comes from the research by Gul et al.~\citep{gul2021structure}. Two problems are addressed in that research and we focus on the first one, detecting toxicity by molecular descriptors. The data set is available in the UCI repository~\citep{kelly2023uci} under the name Toxicity.

The Toxicity dataset was developed to evaluate the toxicity of molecules designed to interact with CRY1, a protein central to the regulation of the circadian rhythm. This biological clock influences numerous physiological processes and its disruption has been associated with diseases such as cancer and metabolic disorders.

The dataset (Table~\ref{datasets}) contains molecular descriptors for 171 molecules obtained by computational calculations, which can be used to train machine learning models capable of predicting whether a molecule is toxic or non-toxic. Each molecule in the dataset is represented by 1203 molecular descriptors, which include physicochemical, topological and structural properties. Examples of descriptors include:
\begin{itemize}
    \item Physicochemical properties: Molecular mass, logP (partition coefficient), number of hydrogen bonds.
    \item Topological descriptors: molecular connectivity indices, number of cycles in the structure.
    \item Electronic properties: Energy of orbitals, electrostatic potential.
\end{itemize}

These descriptors are generated by computational chemistry software and are commonly used in toxicity prediction models.

Non-toxic, the majority class, accounts for 67.25\% of the instances. Therefore, a classifier that always predicts the majority class would achieve an accuracy of 67.25\%. As a result, a good performance from any algorithm should exceed this baseline to demonstrate its effectiveness in distinguishing between classes.

\begin{table}[htbp]
\centering
\begin{tabular}{lrrrrr}

\toprule
 \textit{Features} & \textit{Instances} & \textit{Clases} & \multicolumn{2}{l}{\textit{Class distribution}} \\
  \textit{(Molecular descriptors)}&  \textit{(Molecules)}& & \textit{Non-toxic} & \textit{Toxic} \\ 
\toprule

 1203 & 171 & 2 & 115 & 56 \\

\bottomrule

\end{tabular}
\caption{Dataset properties.}
\label{datasets}
\end{table}

\subsection{Classifiers}
The DTC, RFC, ETC and XGBC classifiers used in~\citep{gul2021structure} and \textit{kNN} were employed for the experimentation.
\textit{kNN}~(k-Nearest Neighbors) is a non-parametric instance-based learning algorithm used for classification and regression tasks. It classifies a new instance by considering the K closest training examples in the feature space, typically measured using Euclidean distance or other distance metrics. The predicted class is determined by a majority vote among the nearest neighbors~\citep{steinbach2009knn,cover1967nearest}. 
DTC (Decision Tree Clasiffier) is a nonparametric supervised learning algorithm, which is used for both classification and regression tasks. It has a hierarchical tree structure, consisting of a root node, branches, internal nodes and leaf nodes~\citep{priyanka2020decision}.
RFC (RandomForest Classifier) is a classifier that relies on combining a large number of uncorrelated and weak decision trees to arrive at a single result~\cite{blanchet2020constructing}. ETC (Extra Trees Classifier) is a classification algorithm based on decision trees, similar to RFC, but with more randomization. Instead of searching for the best splits at each node, it randomly selects features and cutoff values. This makes it faster and less prone to overfitting. It works by creating a set of decision trees and makes predictions by majority vote~\citep{sharaff2019extra}. XGBC (Extreme Gradient Boosting Clasiffier) is a machine learning algorithm based on boosting, which builds sequential decision trees to correct errors in previous trees. It is efficient, fast and avoids overfitting thanks to its regularization. It uses gradient descent to optimize the model and is known for its high classification accuracy. However, it can be complex to fit and has less interpretability than other models~\citep{chen2015xgboost}.

\subsection{Development and Running Environment}
The GA for feature selection has been programmed in Python, using the library DEAP (Distributed Evolutionary Algorithms in Python), an evolutionary computation framework for rapid prototyping and testing of ideas~\citep{fortin2012deap}. It seeks to make algorithms explicit and data structures transparent. Likewise, we have used Scikit-learn~\citep{pedregosa2011scikit} which is a free software machine learning library for the Python programming language. Experiments have been run on a cluster of 6 nodes with Intel Xeon E5420 CPU 2.50GHz processor, under Ubuntu 22.04 GNU/Linux operating system.

\subsection{Experimental parameters}\label{parameters}

The experimental setup initially attempts to mimic the study with which we compare, the original study that defined the data set~\cite{gul2021structure}. The same grid search using train-test 10 fold cross-validation is applied to choose the parameter values from those described in Table~\ref{classifierParameter}. This includes the parameters for \textit{kNN} classifier which is incorporated in our experimentation to introduce the diversity by considering a proximity based classifier. 

\begin{table}[htbp]
\centering
\begin{tabular}{lp{1.5cm}p{2.2cm}p{3cm}p{1.8cm}}
\toprule
\textit{Parameter}\textsuperscript{†} & \textit{kNN} & \textit{DTC}&  \textit{RFC} and  \textit{ETC}&  \textit{XGBC}\\
\midrule
\textbf{k} & [1,3,5,7,9] &-----&-----&----- \\ \midrule
\textbf{max\_depth} &-----& [1, 2, ..., 10, None] & [1, 2, ..., 6, None] & [3, 5, 7, 10] \\ \midrule
\textbf{min\_samples\_splits} &-----& [2, 3, ..., 10] & [2-5] & ----- \\ \midrule
\textbf{min\_samples\_leafs} &-----& [1, 2, ..., 10] & [1-5] & ----- \\ \midrule
\textbf{max\_features} &-----& [1, 2, ..., num\_features] & [1, 2, ..., sqrt(num\_features)] & ----- \\ \midrule
\textbf{n\_estimators} &-----& ----- & [100, 200] & [100, 200] \\ \midrule
\textbf{learning\_rate} &-----& ----- & ----- & [0.01, 0.1] \\ \midrule
\textbf{min\_child\_weight} &-----& ----- & ----- & [1, 3, 5] \\ \midrule
\textbf{subsample} & ----- &-----& ----- & [0.5, 0.7] \\ \midrule
\textbf{colsample\_bytree} &-----& ----- & ----- & [0.5, 0.7] \\ 
\bottomrule
\end{tabular}
\begin{description}
    \setlength{\itemsep}{0pt}  %
    \setlength{\parskip}{0pt}  %
    \item[ \textnormal{\textsuperscript{†}}\hspace{0.3cm}\textbf{k}] The number of nearest neighbours used to classify in \textit{kNN}.
    \item[\hspace{0.5cm}\textbf{max\_depth}] The maximum depth of the tree.
    \item[\hspace{0.5cm}\textbf{min\_samples\_splits}] The minimum number of samples required to split an internal node.
    \item[\hspace{0.5cm}\textbf{min\_samples\_leafs}] The minimum number of samples required to be at a leaf node.
    \item[\hspace{0.5cm}\textbf{max\_features}] The number of features to consider when looking for the best split.
    \item[\hspace{0.5cm}\textbf{n\_estimators}] The number of trees in the forest.
    \item[\hspace{0.5cm}\textbf{learning\_rate}] Step size shrinkage used in update to prevent overfitting.
    \item[\hspace{0.5cm}\textbf{min\_child\_weight}] Minimum sum of instance weight needed in a child.
    \item[\hspace{0.5cm}\textbf{subsample}] Subsample ratio of the training instances.
    \item[\hspace{0.5cm}\textbf{colsample\_bytree}] Subsample ratio of columns when constructing each tree.
\end{description}
\caption{Parameters of the classifiers tested in grid search.}
\label{classifierParameter}
\end{table}

\begin{figure}[htbp]
  \centering
  \includegraphics[width=0.65\textwidth]{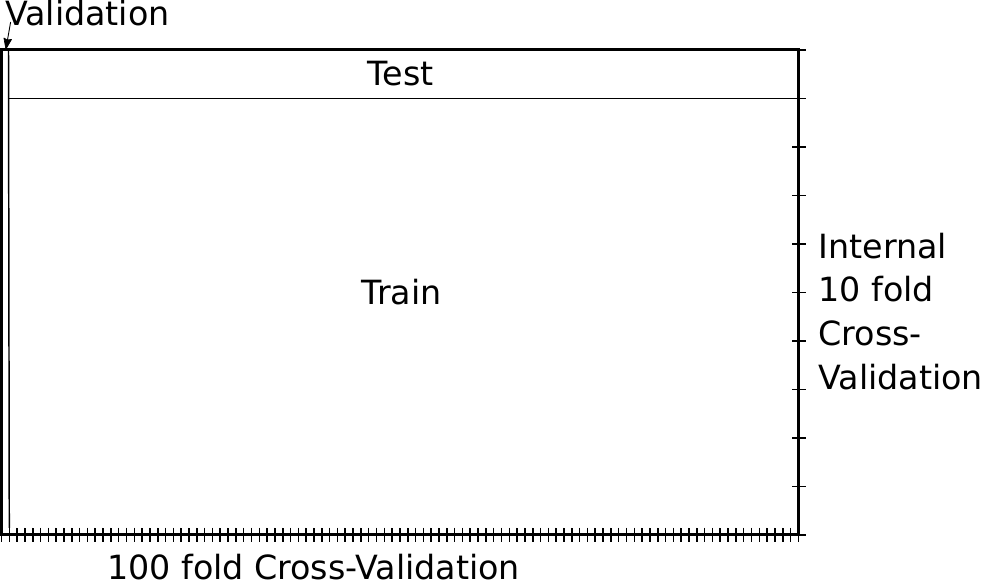}
\caption{Illustration of one iteration of the nested cross-validation.}
\label{fig:validation}
\end{figure}

To evaluate the performance of the classification, the same internal cross-validation of ten partitions from grid search optimization is used in the original study. To obtain stable results they repeated the experimentation 100 times. However, giving that the number of combinations evaluated by grid search is on the order of the ten thousands, over-fitting is likely to occur,  potentially leading to an overestimation of the expected accuracy. To address this issue, instead of repeating the experimentation 100 times, we utilize nested cross-validation with 100 folds. In this way, the results are expected to be similar in testing, as only a very small number of instances are omitted in each run, while providing a reliable estimate of the expected performance. The whole process is repeated 10 times to ensure a stable final validation result. This evaluation is illustrated in figure~\ref{fig:validation}.

After reproducing the experimentation with the 13 features selected in the original study, the GA is applied with the goal of finding a better feature set allowing to fit a more effective classifiers. In order to configure the parameters of the GA, based on previous experimentation, we adopted the values proposed in~\citep{arauzo2025simplecooperative}, which are detailed in Table~\ref{ParameterEvolutive}. Since using grid search inside the wrapper cross validation used as fitness measure is unfeasible, the default parameters from scikit-learn are used for the classifiers and 1 nearest neighbour for \textit{kNN}.

\begin{table}[htbp]
\centering
\begin{tabular}{lll}
\toprule
\textit{Parameter} & \textit{Description} & \textit{Value}\\
\midrule
\textbf{p\_crossover} & Crossover probability & 0.75 \\
\textbf{p\_mutation} & Mutation probability & 0.15 \\
\textbf{elitims} & Number of best individuals kept for next generation & 0\\
\textbf{init\_prob} & Probability initial features & 0.01\\
\textbf{population} & Number of individuals in each population & 50\\
\textbf{generations} & Number of generations to evolve & 300\\
\bottomrule
\end{tabular}
\caption{Parameter values fixed in the genetic algorithm.}
\label{ParameterEvolutive}
\end{table}

In order to find a good setup for feature selection GA, the parameter values shown in table~\ref{ParameterFase1} are tested.

\begin{table}[htbp]
\centering
\begin{tabular}{lll}
\toprule
\textit{Parameter} & \textit{Description} & \textit{Values}\\
\midrule
\textbf{penalty} & Weight for reduction goal in fitness & 0.3, 0.5, 0.7\\
\textbf{var\_penalty} & Variance penalty applied  & Yes, No \\
\bottomrule
\end{tabular}
\caption{Parameter values tested for the genetic algorithm.}
\label{ParameterFase1}
\end{table}

\section{Results analysis}

As an initial step, we replicated the experimental evaluation presented in~\citep{gul2021structure}, using the same 13 selected features and sticking to the same classification processes and software for all classifiers, including the \textit{kNN} classifier. This replication ensures methodological consistency and enables a direct comparative analysis with the original results. The only difference is that we have transformed the 100 repetitions into a second level 100 folds cross-validation by prescinding of one or two instances per repetition. Given the small number of omitted instances, their impact on the results is expected to be negligible, while it establishes a confident approach to evaluate potential improvements and assess the robustness of the models.

The 13 most relevant features were selected using Recursive Feature Elimination (RFE). The classification results obtained with this subset of features are shown in Table~\ref{reproduccion} together with those reported in the original proposal. It is unclear to us the value reported in the original proposal. We believe it must be the average test accuracy of the best model found, not the average of the 100 runs over the average test because that value is closer to the best model column in table~\ref{reproduccion} (which is the best model according to average 10 fold CV test accuracy). With that consideration, the results obtained seem similar, confirming the validity of their experimentation.

\begin{table}[htbp]
\centering
\begin{tabular}{llllll}
\toprule
\textit{Classifier} & \textit{Original~\citep{gul2021structure}} & \textit{Worst} & \textit{Avg. test accuracy$\pm{std}$} & \textit{Best} &\textit{Validation} \\
\toprule
\textit{kNN} & ------ & 0.6276 & 0.6384$\pm{0.0066}$  & 0.6570 & 0.6485\\ \midrule
DTC          & 0.7963 & 0.7217 & 0.7479$\pm{0.0121}$ & 0.7765 & \textcolor{customhighgreen}{0.7392}\\ \midrule
RFC          & 0.7236 & 0.6765 & 0.6921$\pm{0.0077}$ & 0.7158 & 0.6585\\ \midrule
ETC          & 0.6857 & 0.6688 & 0.6764$\pm{0.0064}$ & 0.6941 & 0.6368\\ \midrule
XGBC         & 0.6887 & 0.6688 & 0.6805$\pm{0.0057}$ & 0.7040 & 0.6439\\
\bottomrule
\end{tabular}
\caption{Results from~\citep{gul2021structure} compared with the models found in the experimental reproduction with 100 repetitions of the 10 fold CV using the same 13 features.}
\label{reproduccion}
\end{table}

Nevertheless, it is important to note that validation indicates the model's expected accuracy is around 74\%, which is lower than the reported best test average of 79\% from the original study (and the 78\% from the reproduced equivalent result).

The DTC classifier achieved the highest validation accuracy (73.92\%), demonstrating its effectiveness when trained with the selected features. However, the performance drop observed for the remaining classifiers between testing and validation suggests that the selected characteristics may primarily favour the DTC, potentially limiting its generalization across different models. Furthermore, it is important to note that the performance of the other classifiers does not exceed the baseline of the majority class, indicating that these models provide very poor discriminative power. To highlight this, those models with validation accuracy over the majority rate are coloured in green. These results highlight the need for further analysis of the feature selection and learning process to ensure robustness and generalization.

\begin{figure}[htbp]
  \centering
  \includegraphics[width=1.0\textwidth]{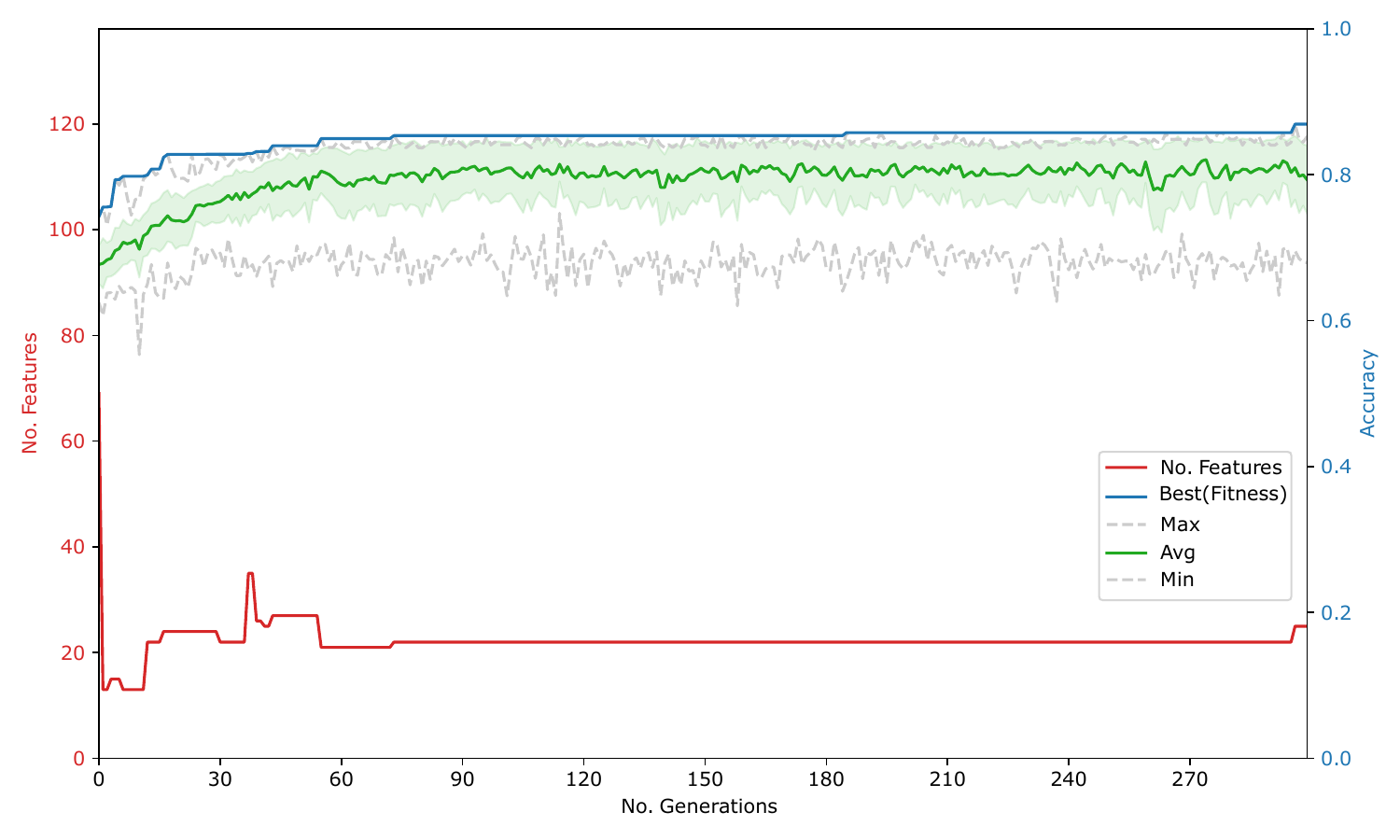}
\caption{Evolution of the GA wrapping DTC with 0.3 feature set penalty and no variance penalty (fourth line of table~\ref{evolutionary}).}
\label{Evolution}
\end{figure}

In the second phase of our study, the GA is used to perform feature selection and evaluate its impact on the different classifiers. As an example, the evolution process of one run of the GA is illustrated in figure~\ref{Evolution}.

The results in Table~\ref{evolutionary} show that the 16 features selected by the GA wrapping RFC achieved the highest validation accuracy (71.87\%) by using a penalty factor of 0.3, demonstrating that the evolutionary approach can improve the generalization capacity of the models studied, as this result is much higher than those achieved for RFC by using RFE in the previous study. In contrast, the DTC classifier, which had previously obtained the highest test accuracy, experienced a significant drop in validation, evidencing a possible overfitting of the training set, probably because it is using too many features.

\begin{table}[htbp]
\centering
\renewcommand{\arraystretch}{1.2}
\begin{tabular}{lrrrrrr}
\toprule
\textit{Classifier} & \textit{noF} & \textit{Penalty} & Min. & \textit{Avg. test accuracy$\pm{std}$} & Max. &\textit{Validation} \\
\toprule
 & 10  & 0.3 & 0.6456 &  0.6596$\pm0.0091$  & 0.6868 & \textcolor{customhighgreen}{0.6760}\\ 
\textit{kNN} & 11  & 0.5 & 0.6269 & 0.6519$\pm0.0091$ & 0.6750 & 0.6006\\ 
 &  8  & 0.7 & 0.6397 & 0.6606$\pm0.0095$ & 0.6868 & 0.6526\\ \midrule
   & 25  & 0.3 & 0.7283 &  0.7478$\pm0.0100$ & 0.7757 & 0.6444\\ 
DTC   & 25  & 0.5 & 0.7224 & 0.7423$\pm0.0099$ & 0.7699 & 0.6550\\ 
   & 12  & 0.7 & 0.7235 & 0.7411$\pm0.0082$ & 0.7588 & 0.6632\\ \midrule
   & 16  & 0.3 & 0.7158 & 0.7359$\pm0.0090$ & 0.7577 & \textcolor{customhighgreen}{0.7187} \\ 
 RFC  &  9  & 0.5 & 0.6989 & 0.7225$\pm0.0085$ & 0.7471 &  \textcolor{customhighgreen}{0.6947}\\ 
   &  7  & 0.7 & 0.7048 & 0.7214$\pm0.0074$ & 0.7460 & \textcolor{customhighgreen}{0.6971}\\ \midrule
   & 18  & 0.3 & 0.6868 & 0.7051$\pm0.0103$ & 0.7279 & \textcolor{customhighgreen}{0.7152} \\ 
 ETC  & 12  & 0.5 & 0.6706 & 0.6902$\pm0.0114$ & 0.7165 & \textcolor{customhighgreen}{0.6766}\\ 
   & 9   & 0.7 & 0.6746 & 0.6905$\pm0.0079$ & 0.7059 & 0.6626\\ \midrule
  & 14  & 0.3 & 0.7044 & 0.7216$\pm0.0089$ & 0.7574 & 0.6714 \\

XGBC  & 9 & 0.5 & 0.7235 & 0.7447$\pm0.0117$ & 0.7813 & \textcolor{customhighgreen}{0.7029} \\

  & 9 & 0.7 & 0.7176 &  0.7356$\pm0.0100$ & 0.7640 & \textcolor{customhighgreen}{0.6988} \\
\bottomrule

\end{tabular}
\caption{Results using features selected by the genetic algorithm.}
\label{evolutionary}
\end{table}

In the third phase, we test the hypothesis that using formula (\ref{eq:fitness-variance}) the  found feature sets are less prone to overfitting. Table~\ref{varpenalty} shows the results of the same experiments of Table~\ref{evolutionary} but using the new fitness calculation with a penalization for non homogeneous accuracy among test partitions. The results seem to improve but the differences are small and they can not be considered conclusive. For this reason we think that more research is needed to find strategies to avoid overfitting in these challenging high dimensional data sets with few instances.

\begin{table}[htbp]
\centering
\renewcommand{\arraystretch}{1.2}
\begin{tabular}{lrrrrrr}
\toprule
\textit{Classifier} & \textit{noF} & \textit{Penalty} & Min. & \textit{Avg. test accuracy$\pm{std}$} & Max. &\textit{Validation} \\
\toprule
 & 10  & 0.3 & 0.6397 & 0.6598$\pm$0.0091 &  0.6868 & 0.6725\\ 
\textit{kNN} &  9   & 0.5 & 0.6393 & 0.6616$\pm$0.0099 & 0.6868 & 0.6673\\ 
 &  4   & 0.7 & 0.6564 & 0.6776$\pm$0.0089& 0.6985 & \textcolor{customhighgreen}{0.7053} \\ 
\midrule
          &  25   & 0.3 & 0.7283 & 0.7593$\pm$0.0119 & 0.7930 & \textcolor{customhighgreen}{0.6895}\\
DTC &  18   & 0.5 & 0.7401 & 0.7662$\pm$0.0091 & 0.7930 & \textcolor{customhighgreen}{0.6994}\\
 &  15  & 0.7 & 0.7404 & 0.7661$\pm$0.0108 & 0.7996 & \textcolor{customhighgreen}{0.7240}\\ \midrule
 &  19  & 0.3 & 0.7099 & 0.7330$\pm$0.0094 & 0.7577 & \textcolor{customhighgreen}{0.7240}\\ 
RFC & 9 & 0.5 & 0.7169& 0.7329$\pm$0.0071 & 0.7522 & \textcolor{customhighgreen}{0.7187}\\ 
 & 7 & 0.7 & 0.6824 & 0.7030$\pm$0.0103 & 0.7279 & \textcolor{customhighgreen}{0.6754} \\ \midrule
 & 37  & 0.3 & 0.6688 & 0.6758$\pm$0.0047 & 0.6941 & 0.6462\\ 
ETC & 13  & 0.5 & 0.6765 & 0.6963$\pm$0.0086 & 0.7221 & \textcolor{customhighgreen}{0.6861} \\ 
 & 11 & 0.7 & 0.6746 & 0.6927$\pm$0.0080 & 0.7099 & \textcolor{customhighgreen}{0.6860}\\ \midrule
 &  34 & 0.3 & 0.7048 & 0.7181$\pm$0.0086 & 0.7518 & \textcolor{customhighgreen}{0.6988} \\ 
XGBC &  13   & 0.5 & 0.6864 & 0.7047$\pm$0.0085 & 0.7335 & \textcolor{customhighgreen}{0.6866}\\ 
 & 9  & 0.7 & 0.6981 & 0.7162$\pm$0.0092 & 0.7529 & \textcolor{customhighgreen}{0.6766}\\ 
\bottomrule

\end{tabular}
\caption{Results using features selected by the genetic algorithm with a fitness penalization based on the variance in CV}
\label{varpenalty}
\end{table}

Table~\ref{tab:comparative} compares the different methods analyzed, including RFE, the GA approach and the variance penalty version of the GA, applied to five classifiers \textit{kNN}, DTC, RFC, ETC and XGBC. The results show that with RFE, DTC achieved the best validation accuracy (73.92\%), separating itself from the other classifiers. However, in the GA and GA with variance penalty  methods, RFC and XGBC showed superior performance, especially with penalties of 0.3 and 0.5. In particular, RFC achieved a remarkable validation performance (71.87\%) with 16 features selected using the GA approach, while XGBC reached its maximum performance (70.29\%) with 9 features and penalty 0.5. However, \textit{kNN} showed inferior performance in most cases, although a higher penalty (0.7) improved its accuracy by selecting only four features. These results suggest that GA and GA with variance penalty methods are effective in improving generalization, especially in ensemble classifiers such as RFC and XGBC, while DTC remains the best classifier with the 13 features selected in the original study.

\begin{table}[htbp]
\begin{center}
\renewcommand{\arraystretch}{1.2}
\setlength{\tabcolsep}{4.5pt}
\begin{tabular}{llrrrrrrrrrr}
\toprule
Type & \textit{P.} & \multicolumn{2}{c}{\textit{KNN}} & \multicolumn{2}{c}{\textit{DTC}} & \multicolumn{2}{c}{\textit{RFC}} & \multicolumn{2}{c}{\textit{ETC}} & \multicolumn{2}{c}{\textit{XGBC}} \\
 & & noF & Val. & noF & Val. & noF & Val. & noF & Val. & noF & Val. \\ \hline
    RFE~\citep{gul2021structure}&---&13&0.6485&13&\textbf{\textcolor{customhighgreen}{0.7392}}&13&0.6585&13&0.6368&13&0.6439\\
                            
    \midrule
    \multirow{3}{*}{Evolutive}&0.3&10&\textcolor{customgreen}{0.6760}&25&0.6444&16&\textcolor{customgreen}{0.7187}&18&\textbf{\textcolor{customhighgreen}{0.7152}}&14&0.6714\\
                              &0.5&11&0.6006&25&0.6550&9&\textcolor{customgreen}{0.6947}&12&\textcolor{customgreen}{0.6766}&9&\textbf{\textcolor{customhighgreen}{0.7020}}\\
                              &0.7&8&0.6526&12&0.6632&7&\textcolor{customgreen}{0.6971}&9&0.6626&9&\textcolor{customgreen}{0.6980}\\

    \midrule
    \multirow{3}{*}{VarPenalty}&0.3&10&0.6725&25&\textcolor{customgreen}{0.6895}&19&\textbf{\textcolor{customhighgreen}{0.7240}}&37&0.6462&34&\textcolor{customgreen}{0.6988}\\
                               &0.5&9&0.6673&18&\textcolor{customgreen}{0.6994}&9&\textcolor{customgreen}{0.7187}&13&\textcolor{customgreen}{0.6861}&13&\textcolor{customgreen}{0.6866}\\
                               &0.7&4&\textbf{\textcolor{customhighgreen}{0.7053}}&15&\textcolor{customgreen}{0.7240}&7&\textcolor{customgreen}{0.6754}&11&\textcolor{customgreen}{0.6860}&9&\textcolor{customgreen}{0.6766}\\

    \bottomrule
\end{tabular}
\end{center}
\caption{Comparison of feature selection for each classifier based on validation results.}
\label{tab:comparative}
\end{table}

\section{Conclusion}

The problem addressed is highly challenging, not only because it belongs to the class of high-dimensional datasets with few instances but also because, after extensive experimentation, achieving a genuine improvement in generalization over the majority class rate with high confidence appears to be highly unlikely.

After reproducing the experiments from the paper that introduced the problem, using an independent validation, we found that the actual expected accuracy is 4\% lower than the reported test accuracy. This highlights the importance of avoiding repeated testing on the same data, as it increases the likelihood of obtaining a fortunate model that performs well under a specific test setup but fails to generalize.

Using the proposed GA to automate the FS process appears promising as, although it has not been able to improve the best model found using DTC in the original study, it has improved the FS performed with RFE for most of the classification models. 

The use of the proposed variance penalty in the GA's fitness function seems promising because it has achieved several better generalization results than the non penalized version. However, it deserves more research to fine-tune it and prove its performance in different data sets.

\funding{This research is supported by projects PID2020-118224RB-I00, PID2023-151336OB-I00 and PID2023-148396NB-I00 financiados por (funded by) MICIU/AEI /10.13039/501100011033, Ministerio de Ciencia e Innovación (Spain, EU).}

\conflictsofinterest{The authors declare no conflicts of interest.}

\abbreviations{Abbreviations}{
The following abbreviations are used in this manuscript:\\

\noindent 
\begin{tabular}{@{}ll}
GA & Genetic Algorithm\\
FS & Feature Selection\\
CV & Cross Validation\\
RFE & Recursive Feature Elimination\\
kNN & k-Nearest Neighbors\\
DTC & Decision Tree Classifier\\
RFC & Random Forest Classifier\\
ETC & Extra Trees Classifier \\
XGBC & Extreme Gradient Boosting Classifier\\
DEAP & Distributed Evolutionary Algorithms in Python\\
\end{tabular}

}

\begin{adjustwidth}{-\extralength}{0cm}

\reftitle{References}

\isAPAandChicago{}{%

}

\isChicagoStyle{%

}{}

\isAPAStyle{%

}{}

\PublishersNote{}
\end{adjustwidth}
\end{document}